# Blind PSF estimation and methods of deconvolution optimization

Yu A Bunyak[1], O Yu Sofina[2] and R N Kvetnyy[2]

[1]InnoVinn Inc. Vinnitsa, Ukraine
[2]Vinnitsa National Technical University, Vinnitsa, Ukraine

E-mail: yuri.bunyak@innovinn.com

**Abstract.** We have shown that the left side null space of the autoregression (AR) matrix operator is the lexicographical presentation of the point spread function (PSF) on condition the AR parameters are common for original and blurred images. The method of inverse PSF evaluation with regularization functional as the function of surface area is offered. The inverse PSF was used for primary image estimation. Two methods of original image estimate optimization were designed basing on maximum entropy generalization of sought and blurred images conditional probability density and regularization. The first method uses balanced variations of convolution and deconvolution transforms to obtaining iterative schema of image optimization. The variations balance was defined by dynamic regularization basing on condition of iteration process convergence. The regularization has dynamic character because depends on current and previous image estimate variations. The second method implements the regularization of deconvolution optimization in curved space with metric defined on image estimate surface. It is basing on target functional invariance to fluctuations of optimal argument value. The given iterative schemas have faster convergence in comparison with known ones, so they can be used for reconstruction of high resolution images series in real time.

## 1. Introduction

Many modern applications need real time reconstruction of high resolution images of some millions pixels size, which are corrupted by defocusing, medium penetration, camera jitter and other factors. Usually, the model of corruption is presented as convolution of original image signal and point spread function (PSF) [5, 23, 27]. The exact PSF shape is unknown in majority cases. So, the image reconstruction problem includes the blind identification of the PSF and deconvolution. The both problems are ill-posed and their solution is approximate in accordance with chosen optimization criterion.

The principal possibility of the PSF estimation problem solution analytically was proven by Lane and Bates [30]. They have shown that convolved components of any multidimensional signal can be separated on condition the signal dimension is greater than one. Their conclusions are based on the analytic properties of the Z-Transform (ZT) in multiple dimensions. The zeros of the ZT of multi dimensional components lie on a hypersurface which can be separated and zeros of the individual components can be recognized up to a complex scale factors. The general approach to zero-sheet separation is based on finding of combinations of some mutually dependent zeros belonging to each of separate ZT coordinate. The required combination creates singularity of a Vandermonde like matrix [1, 30]. This condition points on presence of polynomial substructure in a polynomial structure of higher order. Such solution is equivalent to finding of null space (NS) of mentioned matrix. The zero-sheet separation approach is very complicated for implementation because zeros accuracy evaluation depends on resolution of image spectral presentation which is limited by available combinations number of spectral components and must not be enormously large. The model of image signal can be used instead of the image directly to reduce the problem complexity. As it was shown in [6, 28], the autoregressive and moving average (MA) model enables to separate original image and PSF as parts which correspond to AR and MA models accordingly. The AR model characteristic polynomial contains zeros, exact or fuzzy, that are equal to ZT zeros. But this zeros set includes original image zeros as well PSF zeros because the AR model parameters can be estimated only by means of measured degraded image. An elimination of the PSF zeros influence can be reached by using some number of images which was gotten as different degradations of same original image. As it was shown in [41], if there are at least three differently degraded images then the original image can be reconstructed exactly as right side NS of the operator which acts as convolution of AR parameters and image data vector. The AR model of high order, which size is compatible with image size, was used. It was defined correctly using some blurred images. Therefore the image in the manner of a set of different planes with sharp edges was exactly reconstructed as NS in [41]. The NS may be presented by single eigenvector or by optimized sum of some eigenvectors [42, 43] which correspond to least eigenvalues of the AR model operator. This approach is useful in cases of strongly noised and blurred images. The eigenvector or the eigenvectors sum are the lexicographical presentation of the original image signal matrix and so the NS approach is appropriate to relatively small images reconstruction for a time far from usual period of image frames storing by camera.

When the PSF is known, the problem of deblurring can be considered as inverse filtration of observed image. Methods of inverse filters implementation are direct, mainly in spectral domain, and indirect, with using maximum likelihood, Bayesian and variational approaches. Wiener spectral method [5, 27] lies in a base of majority methods of inverse filtration [11, 37, 40]. The designed methods are intended for regularization of spectrum inversion and optimization of PSF spectrum shape with aim of elimination of noise influence.

There are some approaches to simultaneous iterative deconvolution and PSF estimation. The Lucy-Richardson (LR) maximum likelihood (ML) method [35, 44] of deconvolution was supplemented in [21] by iterative schema of PSF estimation and was developed in [7, 18, 22, 48] in extended manner. The LR method has some limitations, such as slow convergence, sensitivity to noise, image artefacts presence. The methods of variational optimization [13, 14, 20, 38, 51], statistical Bayesian and maximum a posterior models of blur [2, 9, 32, 33, 34, 37, 39, 47] are using to obtain the iterative schemas of sought image and PSF estimates. Limitations and advantages of variational method depend on estimated prior model of PSF and regularization operator. In the case of large image its implementation is difficult because it uses operators in a manner of matrix with rows of size equal to size of lexicographical presentation of processed image. The parametrization of least square, Bayesian or Euler-Lagrange (EL) optimization problems in accordance with gradient methods gives appropriate schemas for iterative optimization of original image estimate [4, 16, 19, 24, 29, 31, 45, 50]. These schemas not always are stable and convergent, they usually require hundreds of iterations for attainment of convergence limit condition.

We consider the problem of arbitrary size images series reconstruction from the point of view of their real time processing. This aim means that original image estimate can be found by one step deconvolution using the inverse point spread function (IPSF) characteristic and it is allowable to use some optimization iterations. It is assumed that blur characteristic changing slowly and IPSF can be estimated without real time restriction. Therefore, the problem solution includes PSF estimation, definition of optimal IPSF and evaluation of primary image estimate, optimization of given estimate by some number of iterations.

## 2. The conjugated NS method of PSF estimation

*2.1. The problem formulation*

The linear model of a degraded measured image $X = X(x, y) \in \Omega$, where $\Omega$ is an image space, $x, y$ – image pixels coordinates, in noise free case has the form of convolution

$$X = H * S \quad (1)$$

of blur characteristic PSF $H \in \Xi \subset \Omega : Power(\Xi) \ll Power(\Omega)$, and original image $S \in \Omega$ [5, 27]. It is required to obtain the inverse expression

$$S = G * X \quad (2)$$

for image reconstruction on condition that the blur function $H$ is unknown. The IPSF is defined in $\Xi$ or it can be defined in $\Omega$ too.

One of the ways of the problem solution is based on the assumption that original and degraded images possess algebraic structure which is invariant to blur operator (1) action. Such common algebraic structure may be defined as operator $\mathrm{A}$ which NS is constituted by the image matrices $X$ and $S$:

$$\int_\Omega \mathrm{A}(x,y;\varsigma,\xi)S(\varsigma,\xi)d\varsigma d\xi \cong 0 ; \qquad (3)$$

$$\int_\Omega \mathrm{A}(x,y;\varsigma,\xi)X(\varsigma,\xi)d\varsigma d\xi \cong 0 . \qquad (4)$$

The transforms (3), (4) are linear and so degradation transform (1) and inverse transform (2) do not influent on their result. The operator $\mathrm{A}$ can be complemented by conjugated null space (CNS) $F \in \Xi$:

$$\int_\Xi F'(\varsigma,\xi)\mathrm{A}(\varsigma,\xi;x,y)d\varsigma d\xi \cong 0 , \qquad (5)$$

where $\bullet'$ – the conjugation. In general, the exact NS does not exist and so we used approximate equality in (3) – (5).

We shall show that matrix convolution transforms, which correspond to expressions (1), (3), (5), on the condition of model operator shift invariance: $\mathrm{A}(x,y;\varsigma,\xi) = \mathrm{A}(x-\varsigma, y-\xi)$, are commutative and can be joined in matrix transform corresponding to (4). Therefore the CNS in (5) can be considered as estimate of the PSF in (1): $F \cong H$.

*2.2. The null space method of image deblurring*

The NS method [41] includes four steps of original image reconstruction.

At the first step the extended matrix is created

$$\mathbf{X} = \left[\mathbf{x}_{i+k,l+m}\right]_{i=1...P, \; l=1...Q}^{k=0...N_x-P-1, \; m=0...N_y-Q-1}, \qquad (6)$$

where low indices point on image matrix columns and high indices point on rows, $N_x \times N_y$ – image matrix $X$ size, $P$, $Q$ – algebraic structure parameters, $\mathbf{x}_{i,k}$ – pixels of matrix $X$, or vectors of pixels of a series matrices like $X$, with indices $i$ and $k$ along coordinates $x$ and $y$.

The second step, the NS vector $\mathbf{a} = \mathbf{vec}(A)$ – the lexicographical presentation of matrix $A = \left[a_{i,k}\right]_{i=1...P}^{k=1...Q}$, has to be found:

$$\mathbf{X} \cdot \mathbf{a} \cong \mathbf{0}. \qquad (7)$$

At the third step the matrix is formed

$$\mathbf{A} = \begin{bmatrix} \mathbf{A}_1 \mathbf{A}_2 ... \mathbf{A}_P \mathbf{0} \; ...\mathbf{0} \\ \mathbf{0} \; \mathbf{A}_1 \mathbf{A}_2 ... \mathbf{A}_P ...\mathbf{0} \\ \vdots \; \; \ddots \; \; \vdots \\ \mathbf{0} \; ...\mathbf{0} \; \mathbf{A}_1 \mathbf{A}_2 ... \mathbf{A}_P \end{bmatrix}; \; \mathbf{A}_k = \begin{bmatrix} a_{k,1} a_{k,2}...a_{k,Q} 0 \; ...0 \\ 0 \; a_{k,1} a_{k,2}...a_{k,Q} ...0 \\ \vdots \; \; \ddots \; \; \vdots \\ 0 \; ...0 \; a_{k,1} a_{k,2}...a_{k,Q} \end{bmatrix}, \qquad (8)$$

where $\mathbf{0}$ – null matrix. Matrix (8) includes $N_x \cdot N_y$ columns, minimal number of rows $P \cdot Q$.

The fourth step, NS vector $\mathbf{s}$ has to be found:

$$\mathbf{A} \cdot \mathbf{s} \cong \mathbf{0}. \qquad (9)$$

The vector

$$\mathbf{s} = \mathbf{vec}(S) \qquad (10)$$

is the lexicographical presentation of the required original image $S$ or its estimate.

*2.3. Null space of PSF*

We can rewrite the expression (7) as the 2D AR model

$$\sum_{i=1}^{P}\sum_{k=1}^{Q}\mathbf{x}_{i+n,k+m}a_{i,k} \approx 0, \tag{11}$$

where $\mathbf{x}_{i,k}$ – image signal samples or RGB vectors, $n = 0...N_x - P$, $m = 0...N_y - Q$. As it was shown in [41], the matrix $\mathbf{X}$ (6) and analogous extended original image matrix $\mathbf{S}$ have the same NS, or in other words, the same AR model (11). This assumption was used in [6, 28] too. We will consider matrix $\mathbf{A}$ (8) as the discrete presentation of the algebraic structure $\mathrm{A}$ in (3) – (5).

As the image (1) is the convolution of original image and PSF, the model (11) is common in respect to these components. But there is one sufficient difference, the images $X, S \in \Omega$, the PSF $H \in \Xi \subset \Omega$. The PSF space $\Xi$ we can present with the help of $\delta$-function window,

$$\Xi = W * \Omega, \ W = W(x,t;y,\tau) = \delta(x-t,\varsigma)\delta(y-\tau,\xi): \ x,y,t,\tau \in \Omega; \ \varsigma,\xi \in \Xi. \tag{12}$$

If we suppose that $H \subset \widetilde{H} : Power\,(\widetilde{H}) = Power\,(\Omega); \ H = W * \widetilde{H}; \ X = \widetilde{H} * S$, then

$$\mathrm{A} * X = \mathrm{A} * \widetilde{H} * S = (W * \widetilde{H})'* \mathrm{A} * S = H'* \mathrm{A} * S = 0. \tag{13}$$

The formulation of rearranging (13) in discrete matrix form has the next manner.

$$\mathbf{A}\left[\mathbf{x}^{(1)}\mathbf{x}^{(2)}...\mathbf{x}^{(K)}\right]^T = \mathbf{A}\left[\widetilde{\mathbf{H}}^{(1)}\widetilde{\mathbf{H}}^{(2)}...\widetilde{\mathbf{H}}^{(K)}\right]^T \mathbf{s} = \left[\mathbf{H}^{(1)}\mathbf{H}^{(2)}...\mathbf{H}^{(K)}\right]\mathbf{A} \cdot \mathbf{s} \cong \mathbf{0}, \tag{14}$$

where $T$ – the transposition, $\mathbf{x}^{(i)} = \mathbf{vec}(X^{(i)})$ – data of $i = 1..K$ degradation filters, $\mathbf{s}$ – the vector (10) of original image, $\widetilde{\mathbf{H}}^{(i)}$ and $\mathbf{H}^{(i)}$ – the $i$-th PSF filter operators with similar structure

$$\mathbf{H} = \begin{bmatrix} \mathbf{H}_1\mathbf{H}_2...\mathbf{H}_L \ \mathbf{0} \ ...\mathbf{0} \\ \mathbf{0} \ \mathbf{H}_1\mathbf{H}_2...\mathbf{H}_L ...\mathbf{0} \\ \vdots \ \ddots \ \vdots \\ \mathbf{0} \ ...\mathbf{0} \ \mathbf{H}_1\mathbf{H}_2...\mathbf{H}_L \end{bmatrix}; \ \mathbf{H}_k = \begin{bmatrix} h_{k,1}h_{k,2}...h_{k,M} \ 0 \ ...0 \\ 0 \ h_{k,1}h_{k,2}...h_{k,M} ...0 \\ \vdots \ \ddots \ \vdots \\ 0 \ ...0 \ h_{k,1}h_{k,2}...h_{k,M} \end{bmatrix} \tag{15}$$

and with different size – the size of the first one $N_x N_y \times N_x N_y$ and the size of the second one $L \cdot M \times P \cdot Q$, $L \times M = Power(\Xi)$. As it follows from matrices (8) and (15) structure, the transform (14) implicitly includes the convolution with space window in (12). The rearranging (13) allows to omit PSF matrices in (14) and to define original image as NS vector (10). The NS method [41] demands absence of coherence between $K$ degradation filters. The filters influence is minimal in equation (7) and it is equivalent to equation (9) at such condition.

In accordance with (9) $\mathbf{A} \cdot \mathbf{s} \cong \mathbf{0}$ in (14), so vector $\mathbf{h} = \mathbf{vec}(H)$ of the operator (15) in (13), (14) can be arbitrary, but on the condition that it is the left side NS,

$$\mathbf{h}^T \cdot \mathbf{A} \cong 0, \tag{16}$$

the vector $\mathbf{h}$ is the PSF because it was defined with the help of splitting of image matrix $X$ and rearranging (13) of its convolved components which are at the symmetric positions in the AR model (11). The vector $\mathbf{h}$ becomes independent with respect to image vector $\mathbf{s}$ as observation characteristic on condition (16). The matrix $\mathbf{A}$ in (16) is sufficiently smaller then same matrix in (9). It includes $L \cdot M$ rows and $(P+L-1) \cdot (Q+M-1)$ columns, $L < P$ and $M < Q$.

With the account of matrix (15) structure, the expression (16) can be rewritten as

$$\mathbf{H} \cdot \mathbf{a} \cong 0, \tag{17}$$

where $\mathbf{a} = \mathbf{vec}(A)$. The PSF and original image can be matched in Z-space as polynomial substructures of general polynomial $A(Z)$ related with AR model (11). The polynomial has local and fuzzy roots which form the zero hypersurface $Z_A : A(Z_A) = 0$. Expressions (7) and (17) are different equations for estimation of the same vector $\mathbf{a}$. So, the ZT of $X$ and $H$ forms hypersurfaces $Z_X \subset Z_A : A(Z_X) = 0$ and $Z_H \subset Z_A : A(Z_H) = 0$. However, $Z_H \subset Z_X$ and $Z_X = Z_S \cup Z_H$. The hypersurfaces $Z_S$ and $Z_H$ are mutually independent (incoherent) – $Z_S \cap Z_H = 0$, because they are defined by ZT in different spaces – $\Omega$ and $\Xi$ correspondingly. And so the hypersurfaces $Z_S$ and $Z_H$ can be separated using the single image fragment. On the other hand, original image $S$ coherent to measured image $X$ and so there were used $K \geq 3$ degraded images with incoherent blur functions for reconstruction of $S$ in [41].

In accordance with expressions (11) and (17) bound elements of the matrix $H$ must vanish because they interact with few number of matrix $\mathbf{A}$ elements which, in general case, are far from zero. Conversely, the central elements of matrix $H$ must neutralize most significant part of matrix $\mathbf{A}$. These conditions determine a convex shape with vanishing bounds of the matrix $H$.

## 3. Deconvolution optimization

### 3.1. Inverse PSF estimation

The IPSF estimation is based on Wiener spectral method [5, 23] which includes division by the PSF spectrum and therefore needs regularization. However, even regularized inverse spectrum gives image boundaries fluctuations. Also, the regularization eliminates filtered image resolution. These deficiencies are caused by discrepancy of image and PSF spaces $\Omega$ and $\Xi$. The artificial matching of PSF into $\Omega$ in spectral transforms by supplementation of zeros creates the spectrum leakage with fluctuations which influence is intensified by division. We shall find the IPSF without exceeding bounds of $\Xi$ to avoid these defects.

Let's write the expressions (1) and (2) in discrete manner.

$$\mathbf{x}_{i,k} = \sum_{l=1}^{L} \sum_{m=1}^{M} \mathbf{s}_{i+l,k+m} h_{l,m}, \qquad (18)$$

$$\mathbf{s}_{i,k} = \sum_{l=1}^{L} \sum_{m=1}^{M} \mathbf{x}_{i+l,k+m} g_{l,m}, \qquad (19)$$

where $\mathbf{x}_{i,k}$ and $\mathbf{s}_{i,k}$ – the samples or RGB vectors of observed distorted image $X$ and sought original image $S$, $h_{l,m}$, $g_{l,m}$ – the elements of PSF and IPSF, $i = 0...N_x - L - 1$, $k = 0...N_y - M - 1$. The substitution (19) into (18) gives the equation for evaluation of the IPSF.

$$\mathbf{x}_{i,k} = \sum_{l=1}^{L} \sum_{m=1}^{M} g_{l,m} \left( \sum_{l'=1}^{L} \sum_{m'=1}^{M} \mathbf{x}_{i+l-l',k+m-m'} h_{l',m'} \right). \qquad (20)$$

The expression in brackets yields once more degradation of observed image by filtering. The indices of $\mathbf{x}_\bullet$ in right part of equation (20) are symmetrized relatively to the left part indices due to symmetry of convolution. The equation (20) is ill-posed because $N_x \gg L$ and $N_y \gg M$. So it needs regularization which applies restrictions on its solution.

We suppose that PSF and IPSF belong to a class of smooth functions. We choose the surface area of IPSF as a criterion of smoothness because the minimal surface area corresponds to the maximally smooth flat surface. The smoothness of the sought IPSF means a minimization of fluctuations caused by image degradation factors. The surface area as regularization functional was used in image denoising problem solution [3, 46]. However, it was not met in formulations of problems of convolutional type with the same aim as in (20).

The surface area of the function $G(x,y)$ is defined as [17]

$$\Sigma_G = \int_\Xi \sqrt{1 + G_x^2 + G_y^2}(x,y)dxdy, \tag{21}$$

where $G_{x(y)} = \partial G(x,y)/\partial x(y)$. The general functional, which includes least squares solution of (20) and regularization term (21), can be written as

$$I_G = \arg\min_G \left[ \frac{1}{2} \int_\Xi \|G*Y - \mathbf{x}\|^2 (x,y)dxdy + \lambda \cdot \Sigma_G \right], \tag{22}$$

where $Y = H*X$ – the matrix which is defined by expression in brackets in (20), $\lambda$ – the regularization parameter.

Euler-Lagrange variational derivative [17] for functional (22) minimum finding looks as

$$\frac{\partial \|G*Y - \mathbf{x}\|^2}{2\partial G} - \lambda \cdot \left( \frac{\partial}{\partial x} \frac{\partial \Sigma_G}{\partial G_x} + \frac{\partial}{\partial y} \frac{\partial \Sigma_G}{\partial G_y} \right) = 0. \tag{23}$$

We also mean fulfillment of zero Neumann boundary condition [17]. With account of (21), the expression (23) takes the next form.

$$Y'*(G*Y - \mathbf{x}) - \lambda \cdot \mathbf{L}_\Sigma(G) = 0, \tag{24}$$

where

$$\mathbf{L}_\Sigma(G) = \left(1 + G_x^2 + G_y^2\right)^{-3/2} \left( \left(1 + G_y^2\right) G_{xx} + \left(1 + G_x^2\right) G_{yy} - 2G_x G_y G_{xy} \right). \tag{25}$$

The discrete iterative schema of equation (24) solution is follows:

$$\mathbf{g}^{(k+1)} = \left(R_{YY} - \lambda \cdot \delta R(\mathbf{g}^{(k)})\right)^{-1} \mathbf{r}_{YX}, \tag{26}$$

where $\mathbf{g} = \mathbf{vec}(G)$, $\mathbf{g}; R_{YY}; \mathbf{r}_{YX} \subset \Xi$,

$$R_{YY} = Y'*Y, \quad \mathbf{r}_{YX} = Y'*\mathbf{x}, \tag{27}$$

$$\delta R(\mathbf{g}^{(k)}) = \left( \mathbf{I} + diag\left[ (D_x \mathbf{g}^{(k)})^2 + (D_y \mathbf{g}^{(k)})^2 \right] \right)^{-3/2} \times$$

$$\left( \left( \mathbf{I} + diag\left[ (D_y \mathbf{g}^{(k)})^2 \right] \right) D_{xx} + \left( \mathbf{I} + diag\left[ (D_x \mathbf{g}^{(k)})^2 \right] \right) D_{yy} - 2 \cdot diag[D_x \mathbf{g}^{(k)}] \cdot diag[D_y \mathbf{g}^{(k)}] D_{xy} \right),$$

$\mathbf{I}$ – the identity matrix, $D$ – the differential operator with pointed by indices variables and degree, $diag[\cdot]$ – diagonal matrix, created by vector in brackets. The parameter $\lambda$ can be chosen as maximal possible value that provides convergence of schema (26) at first $q$ iterations:

$$\left\langle \|\mathbf{g}^{(k+1)} - \mathbf{g}^{(k)}\|^2 \right\rangle \theta \le \left\langle \|\mathbf{g}^{(k)} - \mathbf{g}^{(k-1)}\|^2 \right\rangle, \tag{28}$$

where $k = 0...q-1$, $\theta$ – positive value, $\langle \cdot \rangle$ – the averaging. The iteration process can be stopped if

$$\left\langle \left\| \mathbf{g}^{(k+1)} - \mathbf{g}^{(k)} \right\|^2 \right\rangle < \varepsilon, \tag{29}$$

where $\varepsilon$ – a small value.

The IPSF (26) can be used for primary estimation of sought original image in accordance with the equation (19). Of course, the spectral method can be used too but in this case the division by the IPSF spectrum is not necessary and influence of spectrum leakage is insignificant.

The principal distinction of the problem (22) and similar denoising and deconvolution problems consists in that the arguments are defined in different spaces – the argument of the problem (22) $G \in \Xi$ and argument of the second ones belongs to image space $\Omega$. We project $\Omega \to \Xi$ by convolution operations (27) and in that way create unified problem presentation space.

*3.2. Lucy-Richardson deconvolution schema and its development*
The iterative LR algorithm [35, 44]

$$S^{(k+1)} = S^{(k)} \left( H'^{(k)} * \frac{X}{S^{(k)} * H^{(k)}} \right) \tag{30}$$

gives the solution of deconvolution problem in accordance with criterion of the ML of observed $X$ and sought original images $S$ ($X$, $S$ are separated RGB components of color image). In the expression (30) $S^{(k)}$ and $H^{(k)}$ – k-th image and PSF estimates, $S^{(0)} = X$. The iterative estimation of PSF was introduced in [21] as

$$H^{(k+1)} = H^{(k)} \left( S'^{(k+1)} * \frac{X}{S^{(k+1)} * H^{(k)}} \right) \tag{31}$$

and was used when exact PSF is unknown. The iterations number of LR schema (30)-(31) can amount to some hundreds. This caused design of some methods of iterations (30) acceleration, for example [7, 48].

The LR algorithm is basing on a maximization of conditional probability distribution (PD) of blurred and initial images using Bayes equation. This equation includes division and therefore schema (30) is unstable, since zero neighborhood zones are inherent to images. Efficiency of the LR algorithm depends on image signal PD type too. Consequently, schema (30) demands regularization and statistical generalization. The last request is achievable by assumption of image PD maximum entropy (ME). This property is providing by Gaussian PD. The regularized and generalized iterative LR ME schema was suggested in [15], it looks as

$$S^{(k+1)} = S^{(k)} + \delta t \cdot \left( H'*X - (H'*H)*S^{(k)} + \lambda \cdot \mathbf{L}(S^{(k)}) \right), \tag{32}$$

where $\delta t$ and $\lambda$ – the parameters of relaxation and regularization, $\mathbf{L}(S)$ – the regularization operator which was obtained as EL equation of a regularization functional. The total variation (TV) functional [45]

$$\mathbf{R}_{TV}(S) = \int_\Omega \sqrt{S_x^2 + S_y^2}\, dxdy \tag{33}$$

was used in [15]. Other linear and nonlinear regularization functionals are also using in deconvolution optimization as well as in denoising – Tikhonov, Tikhonov-Miller (TM), TV with additional $\beta$-factor – $\sqrt{S_x^2 + S_y^2 + \beta}$, $0 < \beta << 1$, in the expression (33), surface area and surface energy, mean curvature, image Laplacian, Laplacian in curved space (Beltrami operator) etc. [3, 8, 12-14, 25, 26, 36].

### 3.3. The method of deconvolution optimization by balanced variations and dynamic regularization (BVDR)

The ME generalization of sought and blurred images conditional PD gives the trivial functional [15]

$$I = \frac{1}{2}\int_\Omega \|X - H*S\|^2 (x,y)dxdy, \qquad (34)$$

which can be optimized by using additional constrain defined by regularization functional $\mathbf{R}(S)$. Then deconvolution optimization means the minimization of the functional

$$I_S = \arg\min_S \left[\frac{1}{2}\int_\Omega \left(\|X - H*S\|^2 + \lambda \cdot \mathbf{R}(S)\right)(x,y)dxdy\right], \qquad (35)$$

where $\lambda$ – the regularization parameter, a positive value or function. The EL equation gives for functional (35) the expression

$$H'*(X - H*S) + \lambda \cdot \mathbf{L}(S) = 0, \qquad (36)$$

where $\mathbf{L}(S) = \delta \mathbf{R}(S)/\delta S$ – the variational derivative that is analogous to the second term in (23). We suppose that IPSF $G$ is found by (20) or (26) and convolution $G*H'$ gives the trivial operator. Then the convolution of IPSF with components of (36) gives the next equation:

$$X - H*S + \lambda \cdot G*\mathbf{L}(S) = 0. \qquad (37)$$

We can parameterize the equation (37) and present it as evolutional equation in accordance with gradient descent method [29, 45],

$$S_t = X - H*S + \lambda(t) \cdot G*\mathbf{L}(S), \qquad (38)$$

where $t$ – the evolution parameter. Neumann boundary conditions are assumed. The regularization parameter $\lambda(t)$ can be defined such that equation (38) has to be convergent.

The convergence of iterative problems in image processing was considered in review paper [24] and some other papers [4, 10, 19, 22, 31]. We can define convergence process with the help of $l_1$ norm as $\langle |S_{t+\tau} - S_t| \rangle \leq \langle |S_t - S_{t-\tau}| \rangle : \tau \to 0$, or as parametric derivative

$$\langle |S_t|_t \rangle = -\chi, \qquad (39)$$

where $\chi$ is a positive small value. We can write expression (39) for (38) as following.

$$\langle |-H*S_t + \lambda_t(t) \cdot G*\mathbf{L}(S) + \lambda(t) \cdot G*\mathbf{L}_t(S)| \rangle = \chi. \qquad (40)$$

Since we have found the regularization parameter as a positive function the equation (40) can be rewritten as inequality

$$\lambda_t(t) \cdot \langle |G*\mathbf{L}(S)| \rangle + \lambda(t) \cdot \langle |G*\mathbf{L}_t(S)| \rangle + \langle |-H*S_t| \rangle \geq \chi. \qquad (41)$$

The $\chi$ is arbitrary value and so the expression (41) can be transformed into the differential equation

$$\lambda_t(t) + \lambda(t)\frac{\langle |G*\mathbf{L}_t(S)| \rangle}{\langle |G*\mathbf{L}(S)| \rangle} = \frac{\langle |H*S_t| \rangle - \chi}{\langle |G*\mathbf{L}(S)| \rangle}, \qquad (42)$$

which solution yields the upper bound of $\lambda(t)$:

$$\sup(\lambda(t)) = \lim_{\chi \to 0} \int_0^t \exp\left(-\int_\tau^t \frac{\langle|G*\mathbf{L}_\varsigma(S)|\rangle}{\langle|G*\mathbf{L}(S)|\rangle}(\varsigma)d\varsigma\right) \frac{\langle|H*S_\tau|\rangle - \chi}{\langle|G*\mathbf{L}(S)|\rangle}(\tau)d\tau + \\ \lambda(0)\exp\left(-\int_0^t \frac{\langle|G*\mathbf{L}_\tau(S)|\rangle}{\langle|G*\mathbf{L}(S)|\rangle}(\tau)d\tau\right).$$ 
(43)

The convergence condition (39) in discrete manner means that $\langle|S^{(k+1)} - S^{(k)}|\rangle < \langle|S^{(k)} - S^{(k-1)}|\rangle$. This condition provides the fulfillment of convergence condition with $l_2$ norm of the residuals like (28)

$$\langle\|S^{(k+1)} - S^{(k)}\|^2\rangle\theta \le \langle\|S^{(k)} - S^{(k-1)}\|^2\rangle$$
(44)

as image is a positive function. The formula (43) in discrete form is following,

$$\lambda^{(0)} = \frac{\langle|H*(S^{(0)} - X)|\rangle}{\delta t \langle|G*\mathbf{L}(S^{(0)})|\rangle}\left(\exp\left(\frac{\langle|G*(\mathbf{L}(S^{(0)}) - \mathbf{L}(X))|\rangle}{\delta t \langle|G*\mathbf{L}(S^{(0)})|\rangle}\right) - 1\right)^{-1};$$

$$\lambda^{(k)} = \left(\lambda^{(k-1)} + \frac{\langle|H*(S^{(k)} - S^{(k-1)})|\rangle}{\delta t \langle|G*\mathbf{L}(S^{(k)})|\rangle}\right)\exp\left(-\frac{\langle|G*(\mathbf{L}(S^{(k)}) - \mathbf{L}(S^{(k-1)}))|\rangle}{\delta t \langle|G*\mathbf{L}(S^{(k)})|\rangle}\right),$$
(45)

where $S^{(-1)} = X$ and $S^{(0)} = G*X$. The discrete form of the equation (38) looks as

$$S^{(k+1)} = S^{(k)} + \delta t \cdot \left(X - H*S^{(k)} + \lambda^{(k)} \cdot G*\mathbf{L}(S^{(k)})\right),$$
(46)

where $\delta t$ – the relaxation parameter.

The iterative schema (46), in contrast to known ones like (32), includes convolutions with PSF and IPSF. There are competitive processes of smoothing and peaking of image shape. If we assume that $\lambda_t(t \gg 0) \approx 0$ in (40) then approximate estimate of the regularization parameter will be the next:

$$\lambda^{(k)} \approx \frac{\langle|H*(S^{(k)} - S^{(k-1)})|\rangle}{\langle|G*(\mathbf{L}(S^{(k)}) - \mathbf{L}(S^{(k-1)}))|\rangle}.$$
(47)

The expression (47) shows that the equation (37) constitutes a balance of variations which are caused by convolutions of PSF and IPSF with sought image estimate and regularization term. The first one smooth variation of image surface and the second one sharpen contours since the regularization operator is a function of image derivatives. The dynamic character of the schema (45)-(46) consists in that the regularization parameter depends on current and previous variations of image estimate.

*3.4. The method of deconvolution optimization in curved space (CS)*
The target of optimization of the functional like (34) is to define such $S_{opt}$ that [17]

$$\frac{\delta I(S_{opt})}{\delta S} \equiv 0. \tag{48}$$

In other words, the optimal value of the functional must be an invariant of argument fluctuation. Let's consider the functional (34) in discrete form for $S = S_{opt}$.

$$\lim_{\Delta S_{opt} \to 0} \frac{1}{2} \sum_{i,j} \|X - H * S_{opt}\|^2 P(\Delta S_{opt})(x_i, y_j) = \min, \tag{49}$$

where $P(\Delta S_{opt}) = \Delta x \Delta y$ – the projection of image surface element on the plane $XOY$. The variation of expression (49) with account of (48) yields nonzero term –

$$\lim_{\Delta S_{opt} \to 0} \frac{1}{2} \sum_{i,j} \|X - H * S_{opt}\|^2 \frac{\partial P(\Delta S_{opt})}{\partial \delta S}(x_i, y_j) \neq 0. \tag{50}$$

The condition (48) will be met for each surface element in (50) if it will be included fully, not by its projection. The surface element in CS with induced by the surface metric is an invariant of variational transforms which cause corresponding tensor transforms of coordinates. It has the following manner [17]: $\Delta S_{opt} = \sqrt{\sigma(S_{opt})} \Delta x \Delta y$, where $\sigma(S) = 1 + S_x^2 + S_y^2$ – the determinant of metric tensor in a point of $\Delta S_{opt}$.

The problem of functional (34) optimization with the account of space curvature can be written as

$$I_S = \arg\min_S \left[ \frac{1}{2} \int_\Omega \|X - H * S\|^2 \sqrt{\sigma(S)}(x, y) dx dy \right]. \tag{51}$$

The condition (48) provides the convexity of the functional (51) and so, existence of local minimum. The EL equation gives for (51) the next expression:

$$H'*(X - H * S) - \frac{1}{4\sigma(S)}(X - H * S)^2 \left( \frac{\partial}{\partial x} \frac{\partial \sigma(S)}{\partial S_x} + \frac{\partial}{\partial y} \frac{\partial \sigma(S)}{\partial S_y} \right) = 0. \tag{52}$$

The convolution of (52) with IPSF and its parameterization give the iterative schema for the problem (51) solution.

$$S^{(k+1)} = S^{(k)} + \delta t \cdot \left( X - H * S^{(k)} + G * \left( \Lambda(H, S^{(k)}) \cdot \mathbf{L}_\Sigma(S^{(k)}) \right) \right), \tag{53}$$

where

$$\Lambda(H, S^{(k)}) = \frac{1}{2\sigma(S^{(k)})} \left( X - H * S^{(k)} \right)^2, \tag{54}$$

the operator $\mathbf{L}_\Sigma(S^{(k)})$ is similar to (25). As it follows from the expressions (53) and (54), the iterative process in every point of image surface is assigned by its own parameter – the corresponding element of the regularization surface $\Lambda$. So, we have obtained the schema of total optimization – optimization in each image point, in contrast to schemas of an integral values optimization like (32) and (46).

We can rewrite the expression (53) relative to $\Delta X^{(k)} = X - H * S^{(k)}$ as

$$(\Delta X^{(k)})^2 + \frac{2\sigma(S^{(k)})}{\mathbf{L}_\Sigma(S^{(k)})} \left( \Delta X_H^{(k)} - \frac{\Delta S_H^{(k)}}{\delta t} \right) = 0, \tag{55}$$

where $\Delta X_H^{(k)} = H'*\Delta X^{(k)}$, $\Delta S_H^{(k)} = H'*\Delta S^{(k)}$, $\Delta S^{(k)} = S^{(k+1)} - S^{(k)}$. Let $\Delta X_H^{(k)} \approx \Delta X^{(k)}$ and $\Delta S_H^{(k)} \approx \Delta S^{(k)}$, then expression (55) is the quadratic equation and local minimum of the functional (51) will be met in a neighbourhood of its roots, or, approximately, in vicinity of

$$S^{(opt)} \approx G*\left( X - \sqrt{\frac{2\sigma(S^{(opt)})}{\mathbf{L}_\Sigma(S^{(opt)})}\frac{\Delta S^{(opt)}}{\delta t}} \right), \quad (56)$$

As it follows from (56), the precision of evaluation of the optimal estimate $S_{opt}$ in (48) is conditioned by relaxation parameter in (56), also in (53). The choosing of $\delta t$ has to be made with account that the function $\mathbf{L}_\Sigma(S^{(k)})$ values can belong to zero neighbourhood. So, the point of $S^{(opt)}$ will be stable on condition that corresponding values of $\Delta S^{(k)}/\delta t$ belong to zero neighbourhood too. This condition gives the lower bound of relaxation parameter region of allowable values:

$$\delta t \geq \frac{\left\langle \left| S^{(k+1)} - S^{(k)} \right| \right\rangle}{\left\langle \left| \mathbf{L}_\Sigma(S^{(k)}) \right| \right\rangle}. \quad (57)$$

*Remark.* The functional (51) can not be used for optimization of IPSF instead (22) because in this case the condition of unified problem presentation space is violated, the regularization surface (54) $\Lambda(G,S) \in \Omega$ and operator (25) $\mathbf{L}_\Sigma(G) \in \Xi$ concurrently.

## 4. Methods implementation and test examples

### 4.1. PSF estimation
The PSF estimation includes the following steps:
1. Evaluation of the matrix $A$ using the equation (10) with vector in the right part, which corresponds to the matrix $A$ central element that is assumed as equal to one. The least squares solution of the equation (10) can be optimized similar to procedure (22)-(27) if image is noised. The image presentable part of the size not least $2 \cdot P \cdot Q \times 2 \cdot P \cdot Q$ may be used for evaluation of the matrix $A$ elements if blur is uniform.
2. Creation of the extended matrix $\mathbf{A}$ (8) and singular value decomposition (SVD) of the product $\mathbf{A} \cdot \mathbf{A}^T$.
3. The PSF matrix $H$ of the size $L \times M$ compilation by means of its lexicographical presentation by the vector of a least singular value.
4. Normalisations of the PSF: $\sum_{l=1}^{L}\sum_{m=1}^{M} h_{l,m} = 1$.

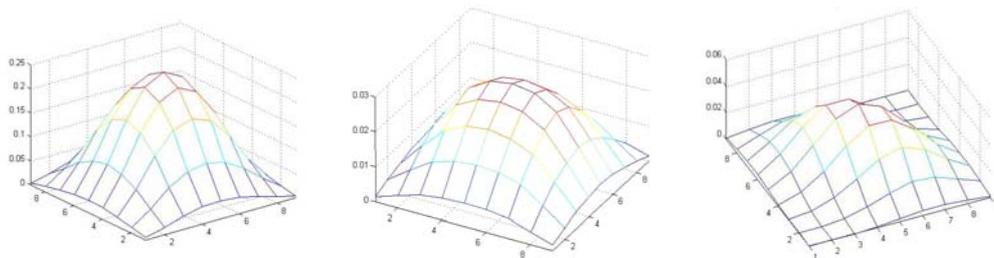

**Figure 1.** Typical estimated PSF shapes of blurs by defocusing and haze (left), horizontal moving (middle) and moving on the angle or vibration (right).

Figure 1 shows the three types of estimated PSF. The first one from the left has symmetrical Gaussian shape and characterizes degradation by defocusing, haze, medium penetration. The next

PSF is Gaussian too but not symmetrical because it characterizes degradation by moving along one of the coordinates, in this case – horizontal moving. The third one corresponds to fast moving on the angle or vibration.

*4.2. IPSF estimation*

We shall consider examples of some images deblurring with different types of blur functions presented in figure 1. The results of deblurring were gotten by convolution with the IPSF. The results of iterative LR deconvolution schema (30)-(31) and of the APEX method [11] are presented too for comparative analysis.

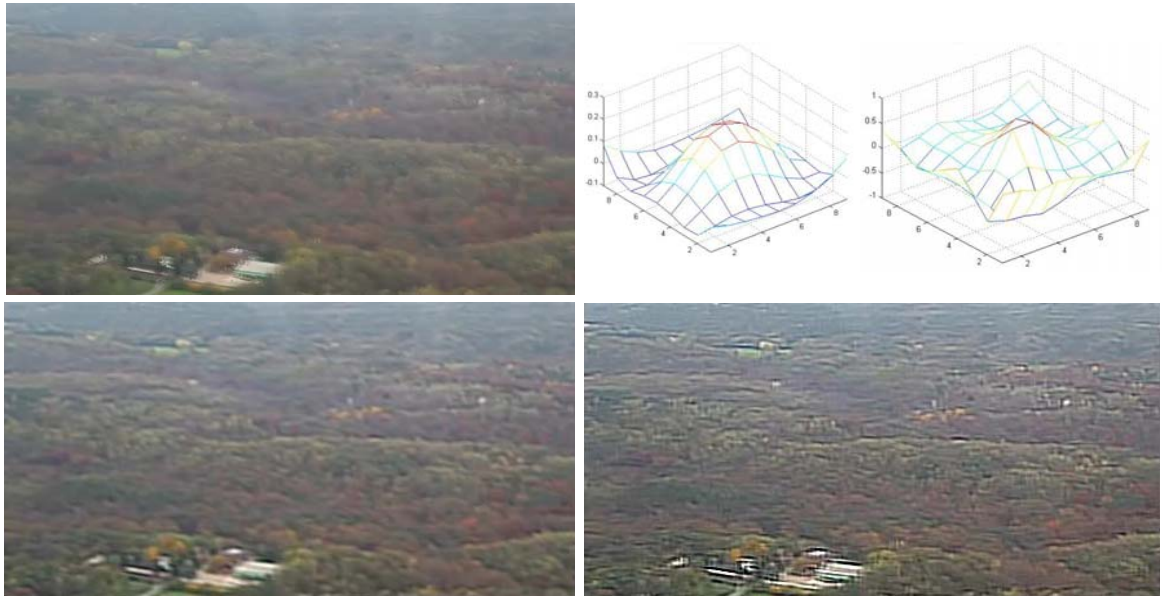

**Figure 2.** Blurred aerial photo (top left), IPSF and optimized IPSF (top center and right), filtered images using IPSF and optimized IPSF (lower left and right correspondingly).

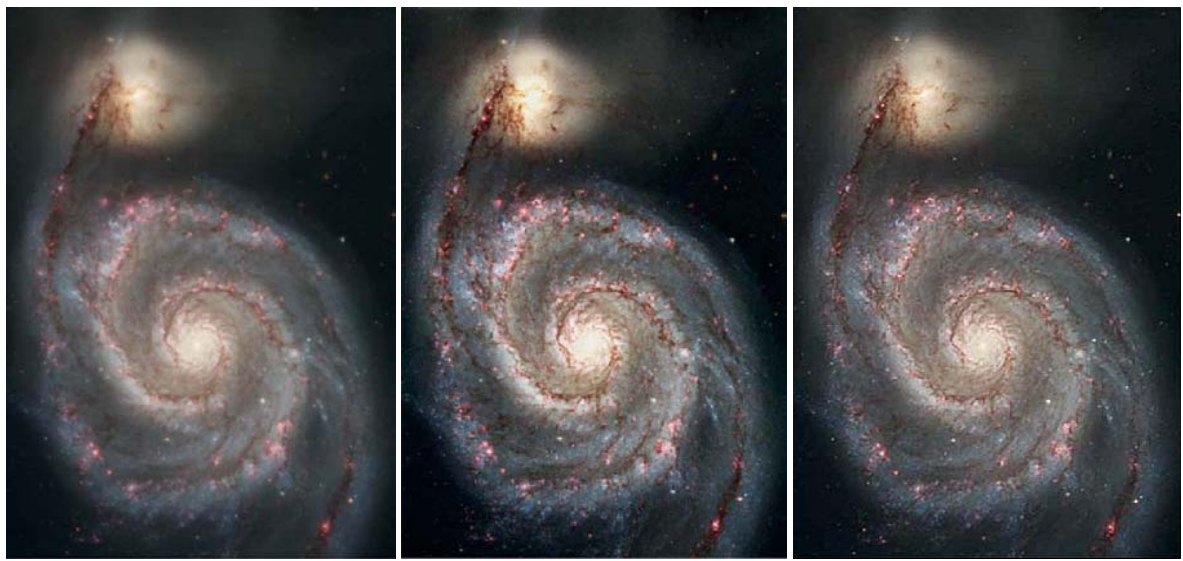

**Figure 3.** The results of one-step deconvolution of astronomical photo (left) by the method [11] (center) and with using the optimized IPSF (right).

Image in figure 2 (top left) is an example of aerial photography blurred by haze and vibration. Its PSF is like to unsymmetrical Gaussian. The AR model of size $P \times Q = 25 \times 25$ was used to evaluate the PSF of size $L \times M = 9 \times 9$. The haze can be considered as noise and so the AR model optimization was used accordingly to schema (22)-(27) with parameters: $\lambda = 0.001$; $q = 3$; $\varepsilon = 1.e-8$. The convergence of the schema (26) was fast, the parameter $\theta$ in (28) amounted up to 10. The next two top images in figure 2 show two IPSF that are found by equations (20) and (26) correspondingly. The optimized IFPS was reached by seven iterations (26). The parameters in expressions (26)-(29) are next: $\lambda = 0.01$; $q = 3$; $\theta \leq 2$; $\varepsilon = 1.e-8$. The obtained two IPSF essentially differ and so results of filtering differ too, they are presented in lower images. It is evident that the convolution with optimized IPSF gives deblurred image (lower right), small details are recognized but contours are excessively sharp and colors are faded.

The first two images in figure 3 was presented in [11] – the first one is camera image, the second one is the result of one-step filtering by the APEX method. The third image was defined by convolution of the first one with the optimized IPSF of size $L \times M = 7 \times 7$, $P \times Q = 17 \times 17$. Obviously, we have obtained the contrast image like previous but without light speckles in the centers of astronomical objects.

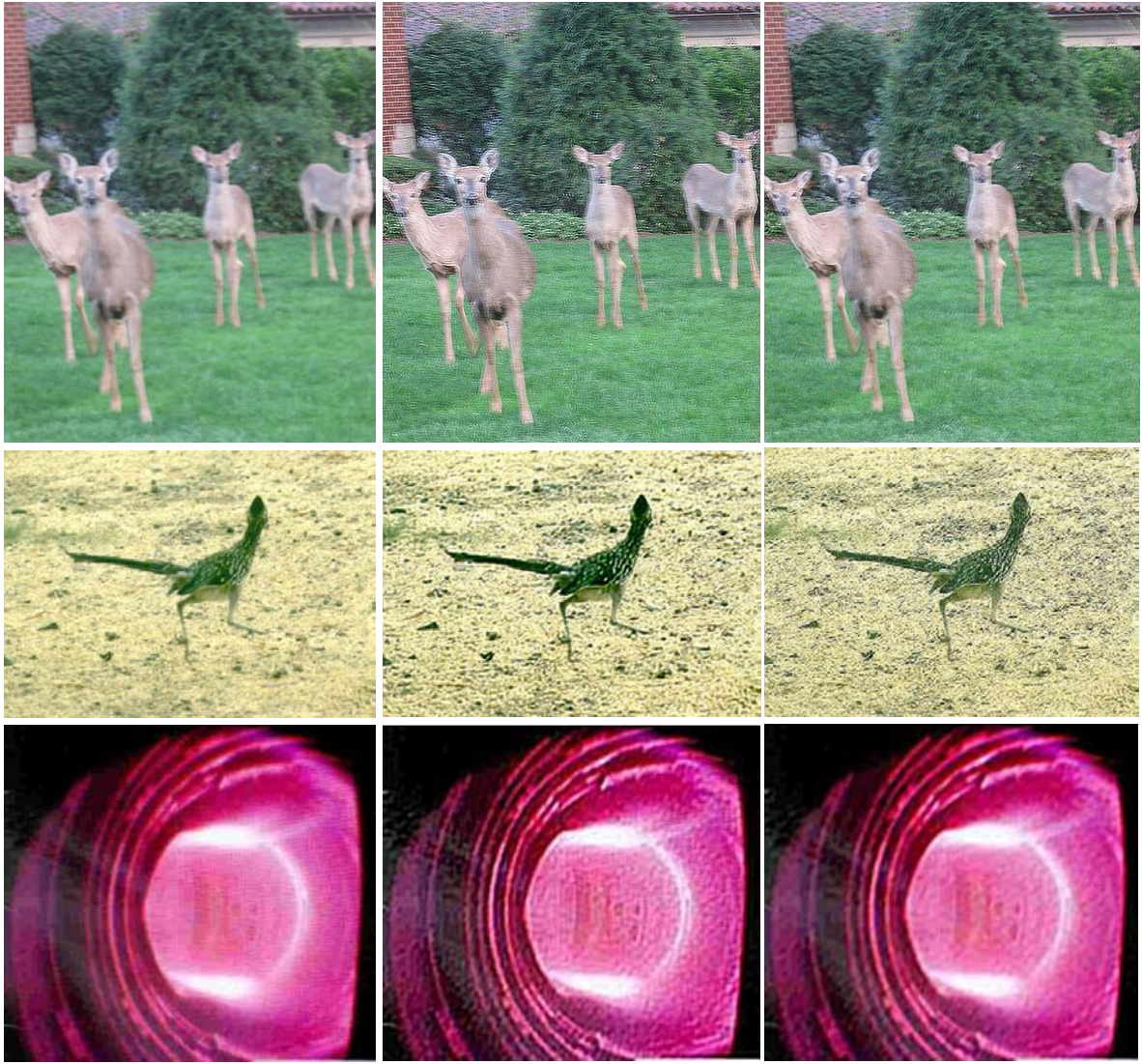

**Figure 4.** Blurred images by moving – left, images reconstructed by LR schema – middle, images filtered by convolution with the IPSF – right.

Figure 4 presents results of deblurring of images corrupted by defocusing and camera moving, camera and object moving, fast moving of gas medium. The original images are unknown but naturalness and small details of reconstructed images can be appreciated. The LR schema (30)-(31) and convolution (19) with the IPSF (26) in spectral domain were used. In the cases of images in figure 4 in lower row the search of regularization parameter which provides the convergence condition (28) at first $q = 3$ steps in the region $0.01 \leq \lambda \leq 0.0001$ was unsuccessful. So, the result of least square solution of the equation (20) with the help of SVD was used.

The order of AR model (11) $P \times Q = 17 \times 17$ and size of the PSF $L \times M = 7 \times 7$ for images in middle and lower rows in figure 4, $P \times Q = 25 \times 25$, $L \times M = 9 \times 9$ for top image in figure 4. The parameters of AR model and IPSF optimization are the same as presented above. The PSF of the first type in figure 1 corresponds to the blurred images in figure 3, the PSF of the second type corresponds to the top and middle images in figure 4 and the PSF of the third type corresponds to the lower image in figure 4.

The estimated by the CNS method PSF (16) was used in LR schema (30)-(31) as initial one. The convergence process of LR schema (30) was stopped at achievement of 100 iteration steps in the cases of two top images in figure 4. It was stopped at violation of the convergence condition (44) after 41 iteration steps in the case of lower image in figure 4. The relative change of the PSF shape by 100 consecutive transforms (31) was less then 0.1%. Opposite, the images shape change by transforms (30) was up to 5%. This fact confirms that the equation (16) gives the estimate of PSF. The images obtained by LR method have clear contours but they contain artefacts which are inherent for this method – colors mosaic and corrupted bounds.

We can see that it is enough of single convolution transform with optimized IPSF for blur removing. The result of convolution with optimized IPSF is reachable by LR method using not less than 10 iterations (30). So, we have single convolution transform opposite to more then twenty ones.

Numerical experiments have shown that if the least square solution of the equation (20) yields estimate of IPSF near to optimal one then its further optimization is not possible because the searching of regularization parameter in accordance with convergence condition (28) is unsuccessful $(\theta = 1)$, an appropriate $\lambda \approx 0$.

The images in figures 2, 4 need colors reconstruction. And so they can serve as primary estimates for further optimization with the aim of contours clearing and colors reconstruction.

*4.3. The examples of deconvolution optimization*

The primary estimates of images in the figure 4 were optimized by the schemas LR ME (32), BVDR (45)-(46) and CS (53)-(54). We used the dynamic regularization approach (39)-(42) to regularization parameter estimation in (32) with the aim to investigate the role of variations balancing in (46). The surface area functional (SAF) (21) was used as regularization term in schemas (32) and (46) because its minimization means maximal possible flatness of image contours neighborhood and, consequently, sharpness of contours.

The numerical experiments have shown that in all examples the schema (32) reaches the condition of convergence termination

$$\left\langle \left\| S^{(k+1)} - S^{(k)} \right\|^2 \right\rangle < 10^{-8} \qquad (58)$$

in 2-3 iteration steps without sufficient influence on the primary estimate $S^{(0)} = G * X$. This happens because the schema (32) contains only smoothing convolution transforms. The results of optimization are presented in figure 6 (left column).

The results of image estimate optimization by the BVDR and CS schemas are shown in figure 5. In the both cases iterative process was restricted by 10 first iteration steps, violation of the condition (44) and the condition (58). The optimization procedures (46) and (53) change colors drawing near them to natural and clearing contours. Different textures and small details show quality of colors tints and contours reconstruction. The BVDR schema is characterized by more sharp contours in comparison with the CS schema at same value of relaxation parameter $\delta t$. The CS schema gives tints closest to blurred image. This is important in video basing measurements, for example, in aggressive medium as plasma in figures 4, 5 (lower rows). Fast moving of the gas medium hides texture, discontinuities and corrupts colors which relate with the turbulence, density and temperature. These features are important for control processes [49].

The process of convergence of the schemas (46) and (53) is different.

The schema (46) has some transition period at the first iteration steps. Initially $\lambda$ (45) grows up to the some maximum value and then monotonically reduces. The residual in (58) repeats $\lambda$ dynamic with one step lag. The transition period duration amounts up to 10 iteration steps depending on image structure and relaxation parameter. If image has many sharp contours then period is long, if image is mainly smooth than the period amounts to 2-4 iteration steps. The transition period amounts to 20 steps and more when relaxation parameter $\delta t > 0.1$. So, we used $\delta t = 0.1$ in estimation of images in figure 5. The transition period did not exceed 5 iteration steps and convergence process was terminated by iteration steps number restriction. The convergence of the schema (46) is conditioned by $\lambda^{(k+1)} \leq \lambda^{(k)}$ ($k >> 1$) in (45) since the $l_1$ norm of smoothed variation in numerator of expression (45) terms is less then the same norm of the excited variation of derivatives function in denominator. The reducing of the regularization parameter causes decrease of the residual in (58). The transition period presence is caused by initial difference between $X$ and $S^{(0)}$ and by dynamic nature of the regularization parameter (45).

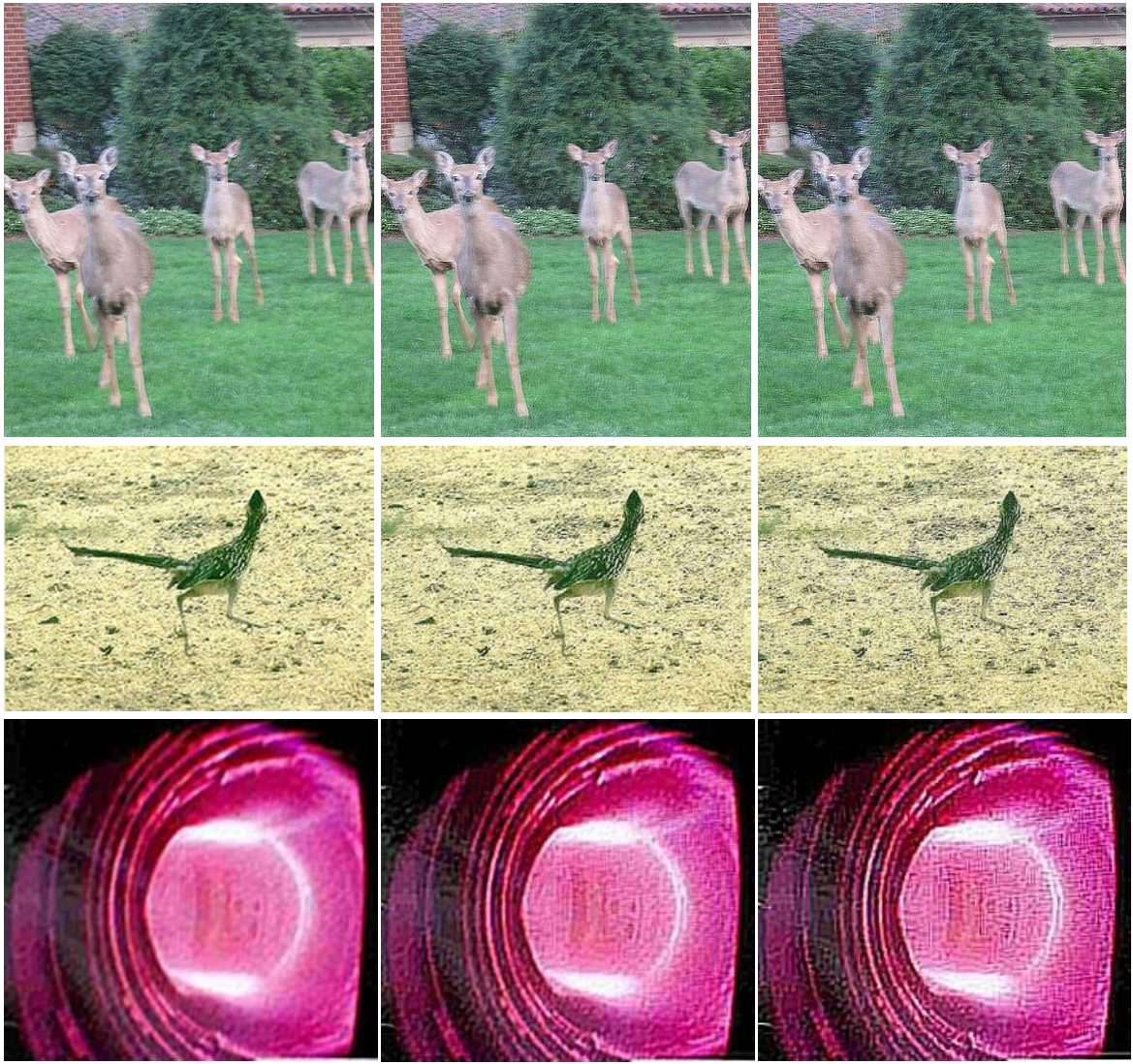

**Figure 5.** Images that are given by optimization of primary estimate (figure 4, at the right) by using LR ME (32), the BVDR (46) and the CS (53) iterative schemas (from the left to the right correspondingly).

The convergence of the schema (53) is monotonous but not unlimited. It reaches some lowest point of the residual in (58) and then the convergence condition (44) is violated. The point of convergence termination in accordance with (56) relates with image structure, filters order and relaxation parameter. For example, in the case of image in lower row in figure 5 the condition (44) violated after 27-th iteration step at $\delta t = 0.1$, after 9-th step at $\delta t = 0.5$ and after 6-th step at $\delta t = 1.0$. The result of the last case is shown in figure 5. The number of iteration steps did not exceed 6 ones for rest examples in figure 5 at $\delta t = 1.0$.

The relaxation parameter restriction (57) was not violated in all examples. As it constitutes lower bound, the ratio

$$\delta t \cdot \max \left| S^{(k+1)} - S^{(k)} \right| \leq 0.01 \qquad (59)$$

was used as higher bound of the relaxation parameter for avoidance of image degradation.

The parameter of convergence velocity $\theta$ did not exceed 1.0001 in (44) for both schemas (46) and (53).

The SAF (21) provides better convergence of iterative schema (46) in comparison with TM, TV and TV with $\beta$ - factor functionals. The schema (46) with mentioned regularization functionals is stable and convergent at $\delta t \ll 0.1$. Therefore, these functionals do not allow to change primary image estimate by some iteration steps essentially.

*4.4. Deblurring of noisy images*

Iterative schemas (46) and (53) are not appropriate for estimation of blurred and noised images because they sharpen all image surface curves and so reinforce noise influence, especially, when noise is impulsive. It is necessary to use a prior image denoised estimate $X_{pr}$ instead $X$ in (46) and (53). Denoised estimate can be obtained by filtering with regularization in spectral region [5, 11, 27, 37, 40] or by the iterative schemas as special procedure [8, 14, 24, 25, 36, 45] or as procedure joined with deconvolution [3, 4, 24, 46]. The filtration based approach introduces an additional smoothing or blur. The iterative schemas need a large number of iteration steps because usually use small value of relaxation parameter.

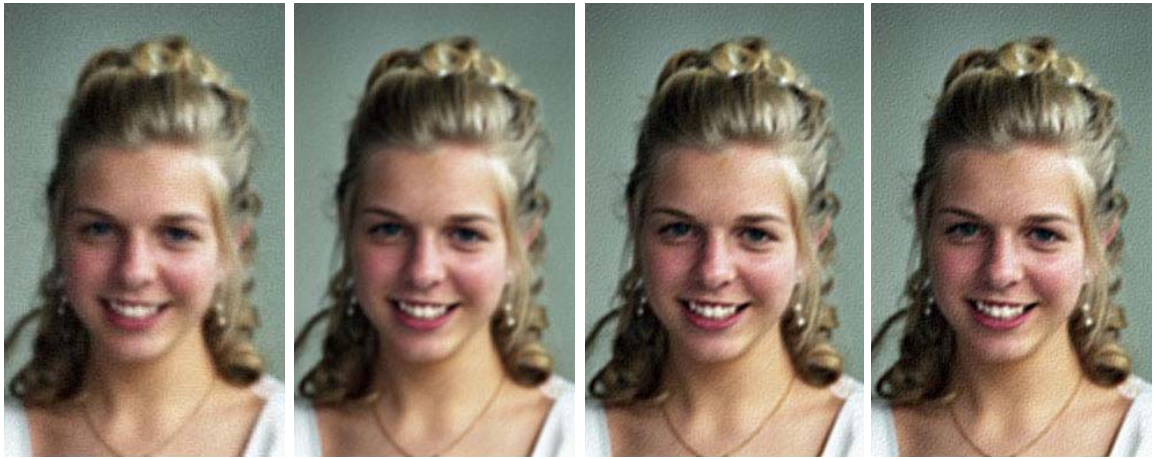

**Figure 6.** Image blurred and corrupted by impulsive noise (first from the left), denoised image by the prior IPSF, primary image estimate by the optimized IPSF, optimized estimate by 20 iteration steps of the BVDR schema.

The filter (19) with transient characteristic which is defined by equation (20) can be used as denoising filter for prior image $X_{pr}$ estimation. This filter bounds initial $X$ and smoothed $H*X$ images. As it can be seen in figure 2, such filter does not introduce additional smoothing. At the same time, it weekly sharpens image surface in comparison with the filter with optimized transient characteristic (26). If the filter order is high then it can eliminates noise influence by weighted accumulation of image matrix elements. The example of blurred and corrupted by impulsive noise image is shown in figure 6. The PSF and IPSF $G_{pr}$ were defined by equations (16) and (20), $P \times Q = 33 \times 33$, $L \times M = 17 \times 17$. The PSF was of the second type in figure 1. Then, the PSF and the optimized IPSF (26) were defined using filtered image $X_{pr} = G_{pr} * X$ (the second one from the left in figure 6). The PSF was of the first type in this case. Its parameters are same as in the prior one. There were obtained primary (the third one in figure 6) and two similar optimized image estimates by the BVDR (at the right in figure 6) and CS schemas. There were used 20 iteration steps with the relaxation parameter $\delta t \sim 0.05$ that was restricted by the ratio (59). The result of optimization of the image estimate without prior filtration was deblurred but hardly corrupted by noise. The second and third images in figure 6 show that the suggested denoising filter can be cascaded with the aim of noise influence elimination. The image was not destroyed by the filters cascade because each next filter is adapted to previous filtering result. The noise pulses of some pixels size did not spread in to spots as it can be in a result of usual smoothing.

**Conclusion**

The AR model is very well studied but its presentation in the manner of convolution operator **A** (8) opens its new ability to signals splitting. The AR model carries the physical sense of passing wave compensation by own reflections. The direct AR operator compensates periodical processes and the conjugated operator compensates impulse processes. If a signal is a convolving of mentioned two types of signals then these signals can be separated because they have different structure and spectrums.

Numerical experiments have shown:
- The iterative transform (31) essentially changes PSF chosen heuristically. The estimated as conjugated null space PSF is an invariant of the transform (31). And so we have obtained (quasi) optimal PSF estimate in accordance with the ML criterion.
- The CNS method is appropriate to PSF estimation when blur is characterizing by a smooth convex function. Sharp image displacements are not available to estimate.
- The PSF size must be agreed with blur level. If the PSF model order is less necessary one then this deficiency can be compensated by additional optimization iterations. If the order is sufficiently greater then required one a restored image will be corrupted. In this case the IPSF has more then one dished maximums.
- The problem (26) has better convergence with regularization functional (21) in comparison with TV functional (33) and others known functionals with derivatives of same order.
- The CS schema (53) is most appropriate to optimization of primary estimate in real time implementation. The required iteration steps for reaching of the functional (51) local minimum does not exceed 10 ones at $\delta t = 1$. As figure 6 has shown, there were obtained similar results by the CS and BVDR schemas with using 5-6 and 10 iteration steps correspondingly. But only metric tensor determinant is applicable for image regularization in the case of CS schema. At the same time, all known regularization functionals can be used in the BVDR schema (45)-(46) with the aim of restriction of image gradient or fluctuation, noise elimination etc.

The suggested deblurring schemas reduce number of requiring iterations more then 10 times in comparison with LR schema when image series is processing and PSF changes slowly in time. No more then twenty convolution procedures are sufficiently for optimized image or its RGB component estimate finding. In many applications image colors reconstruction is not required and then one-step convolution transform is sufficiently. The image reconstruction can be implemented in real time by multi core processors array or graphic processors units at such conditions.

**References**

[1] Aogaki S, Moritani I, Sugai T, Takeutchi F and Toyama F M 2006 Simple method to eliminate blur based on Lane and Bates algorithm *Preprint* cs/0609165
[2] Babacan S D, Molina R and Katsaggelos A K 2009 Variational Bayesian blind deconvolution using a total variation prior *IEEE Trans. Image Process.* **18** 12–26
[3] Bar L, Brook A, Sochen N and Kiryati N 2007 Color image deblurring with Impulsive Noise *IEEE Trans. Image Proc.* **16** 1101–11
[4] Bar L, Sochen N and Kiryati N 2007 Convergence of an iterative method for variational deconvolution and impulse noise removal, *SIAM Multiscale Modeling and Simulation* **6** 983–94
[5] Biemond J, Lagedijk R L and Mersereau R M 1990 Iterative methods for image deblurring *Proc. IEEE* **78** 856–83
[6] Biemond J, Van der Putten F G and Woods J W 1988 Identification and restoration of images with symmetric noncausal blurs *IEEE Trans. Circuits and Systems* **35** 385–93
[7] Biggs D S C and Andrews M 1997 Acceleration of iterative image restoration algorithms *App. Opt.* **36** 1766–75
[8] Brito-Loeza C and Chen Ke 2010 Multigrid algorithm for high order denoising *SIAM J. Imag. Sc.* **3** 363–89
[9] Bronstein M M, Bronstein A M, Zibulevsky M and Zeevi Y Y 2005 Blind deconvolution of images using optimal sparse presentations *IEEE Trans. Image Process.* **14** 726–36
[10] Cai J F, Osher S and Shen Z 2009 Convergence of the linearized Bregman iteration for ℓ1-norm minimization *Math. Comput.* **78** 2127–36
[11] Carasso A S 2006 APEX blind deconvolution of color Hubble space telescope imagery and other astronomical data *Opt. Eng.* **45** 107004 1–15
[12] Chambolle A 2004 An algorithm for total variation minimization and application *J. of Math. Imaging and Vis.* **20** 89–97
[13] Chan T F and Wong C 1998 Total variation blind deconvolution *EEE Trans. Image Proc.* **7** 370–75

[14] Chan T F, Esedoglu S, Park F and Yip A 2005 Recent developments in total variation image restoration *The Handbook of Mathematical Models in Computer Vision* Ed. by: N Paragios, Y Chen and O Faugeras (New York: Springer)
[15] Dey N et al 2004 3D microscopy deconvolution using Richardson-Lucy algorithm with total variation regularization *INRIA preprint* 5272
[16] Donatelli M., Estatico C, Martinelli A and Serra-Capizzano S 2006 Improved image deblurring with anti-reflective boundary conditions and re-blurring *Inverse Problems* **22** 2035–53
[17] Dubrovin B A, Fomenko A T and Novikov S P 1992 *Modern Geometry – Methods and Applications Pt.1* (New York: Springer)
[18] Fish D A, Brinicombe A M, Pike E R and Walker J G 1995 Blind deconvolution by means of the Richardson–Lucy algorithm *J. Opt. Soc. Am.* **A 12** 58–65
[19] Han J, Han L, Neumann M and Prasad U 2009 On the rate of convergence of the image space reconstruction algorithm *Operators and matrices* **3**(1) 41–58
[20] Hao Z, Yu L and Qinzhang W 2007 Blind image deconvolution subject to bandwidth and total variation constraints *Optics Letters* **32** 2550–52
[21] Holmes T J 1992 Blind deconvolution of quantum-limited incoherent imagery: maximum likelihood approach *J. Opt. Soc.Am.* **A 9** 1052–61
[22] Jiang M and Wang G 2003 Convergence studies on iterative algorithms for image reconstruction *IEEE Trans. Med. Imaging* **22** 569–79
[23] Jiang M and Wang G 2003 Development of blind image deconvolution and its applications *J. of X-Ray Sc. and Tech.* **11** 13–19
[24] Katsaggelos A K, Babacan S D and Tsai C-J 2009 Iterative image restoration *The Essential Guide to Image Processing* Ed. by A Bovik (Elsevier)
[25] Keren D and Gotlib A 1998 Denoising color images using regularization and correlation terms *J. of Vis. Comm. and Image Repr.* **9**(4) 352–65
[26] Kimmel R, Malladi R and Sochen N 2000 Images as embedded maps and minimal surfaces: movies, color, texture, and volumetric medical images *Int. J. of Computer Vision* **39**(2) 111–29
[27] Kundur D and Hatzinakos D 1996 Blind image deconvolution *IEEE Signal Proc. Mag.* No 5 43–64
[28] Lagendijk R L, Tekalp A M and Biemond J 1990 Maximum likelihood image and blur identification: a unifying approach *Opt. Eng.* **29** 422–35
[29] Landweber L 1951 An iterative formula for Fredholm integral equations of the first kind *Am. J. Math.* **73** 615–24
[30] Lane R G and Bates R H T 1987 Automatic multidimensional deconvolution *J. Opt. Soc. Am. A* **4(1)** 180–88
[31] Liang L and Xu Y 2003 Adaptive Landweber method to deblur images *IEEE Signal Process. Lett.* **10**(5) 129–32
[32] Liao H; Ng M K 2011 Blind Deconvolution using generalized cross-validation approach to regularization parameter estimation *IEEE Trans. Image Process.* **20** 670–80
[33] Levin A, Weiss Y, Durand F and Freeman W T 2011 Understanding blind deconvolution algorithms *IEEE Trans. Patt. Anal. and Machine Intel.* **33** 2354-67
[34] Likas A C and Galatsanos N P 2004 A variational approach for Bayesian blind image deconvolution *IEEE Trans. Signal Process.* **52** 2222-33
[35] Lucy L B 1974 An iterative technique for rectification of observed distributions *Astronom. J.* **79** 745–54
[36] Lysaker M, Osher S and Xue-Cheng Tai 2004 Noise removal using smoothed normals and surface fitting *IEEE Trans. Image Process.* **13** 1345-57
[37] Michailovich O V, Tannenbaum A 2007 Blind deconvolution of medical ultrasound images: a parametric inverse filtering approach *IEEE Trans. Image Process.* **16** 3005-19
[38] Molina R, Mateos J and Katsaggelos A 2006 Blind deconvolution using a variational approach to parameter image and blur estimation *IEEE Trans. Image* Process **15** 3715-27
[39] Money J H and S H Kang 2008 Total variation minimizing blind deconvolution with shock filter reference *Image and Vision Computing* **26**(2) 302-14
[40] Ng M, Chan R and Tang W 2000 A fast algorithm for deblurring models with Neumann boundary conditions *SIAM J. Sci. Comput.* **21** 851–66
[41] Pai H-T and Bovik A C 1997 Exact multi-channel blind image restoration *IEEE Signal Process. Lett.* **4** 217–20
[42] Pai H-T, Bovik A C and Evance B L 1997 Multi-channel blind image restoration *Turkish J. of Electrical Eng. & Computer Sciences* **3** 79–97
[43] Pai H-T and Bovik A C 2001 On eigenstructure-based direct multichannel blind image restoration *IEEE Trans. Image Process.* **10** 1434–46
[44] Richardson W H 1972 Bayesian-based iterative method of image restoration *J of the Opt. Soc. of Am.* **62** 55–59
[45] Rudin L I, Osher S and Fatemi E 1992 Nonlinear total variation based noise removal algorithms *Physica* **D 60** 259–68
[46] Tschumperle D and Deriche R 2005 Vector-valued image regularization with PDE's: a common framework for different applications *IEEE Trans. Patt. Anal. and Machine Intel.* **27**(4) 506–17
[47] Tzikas D G, Likas A C and Galatsanos N P 2009 Variational Bayesian sparse kernel-based blind image deconvolution with Student's-t priors *IEEE Trans. Image Process.* **18** 753–64
[48] Singh M K, Tiwary U S and Kim Y H 2008 An adaptively accelerated Lucy-Richardson method for image deblurring *EURASIP J. on Adv. in Signal Process.* **2008** 365021
[49] Wagner F 2007 A quarter-century of H-mode studies *Plasma Phys. Control. Fusion* **49** B1–B33
[50] Yang, W Q, Spink D M, York T A and McCann H 1999. An image reconstruction algorithm based on Landweber iteration method for electrical capacitance tomography *Meas. Sci. Technol.* **10** 1065–69
[51] You Y-L, Kaveh M 1996 A regularization approach to joint blur identification and image restoration *IEEE Trans. Image Process.* **5** 416–28